\documentclass{article}
\usepackage{spconf,amsmath,amssymb,graphicx}
\usepackage{booktabs,multirow,array}
\usepackage{float}

\graphicspath{{figures/}}

\makeatletter
\def\title#1{\gdef\@title{#1}}
\makeatother

\title{Segmentation-Guided Spatial Indexing for Generalizable
and Explainable Deepfake Detection}

\name{Izaldein Al-Zyoud\sthanks{Corresponding author: \texttt{izzy.alzyoud@uottawa.ca}} \qquad Abdulmotaleb El~Saddik}
\address{MCRLab, School of Electrical Engineering and Computer Science\\
University of Ottawa, Ottawa, ON, Canada}

\begin{document}
\ninept
\maketitle

\begin{abstract}
We introduce segmentation-guided spatial indexing for generalizable
and explainable deepfake detection.
The key idea reverses the standard design order: rather than pooling
all facial tokens and classifying afterward, we first select
semantically meaningful patch tokens, then pool only those.
A frozen FaRL parser assigns each DINOv3 ViT-L/16 patch token a
semantic label; non-target tokens are discarded; a linear probe
classifies the retained region.
This spatial indexing exploits DINOv3's patch-level spatial
consistency, the same property that enables emergent segmentation,
to present the probe with a purer regional subspace where
manipulation-relevant evidence is less diluted by whole-face cues.
Region attribution is structural: when the mouth model predicts fake,
the decision used only mouth tokens, not an overlaid saliency map.
On Celeb-DF v2, the mouth-indexed probe achieves AUC 0.905,
outperforming LipForensics ($+$8.1\,pp) and Xception ($+$16.9\,pp),
with no DINOv3 or FaRL fine-tuning and no target-domain data.
Ablations isolate the mechanism: replacing regional selection with
DINOv3's CLS token drops Celeb-DF v2 AUC by 26.4\,pp; replacing
DINOv3 with FaRL features drops it by 20.9\,pp.
Both DINOv3 representation and the spatial index are independently
necessary; neither alone approaches the full system.
\end{abstract}


\section{Introduction}
\label{sec:intro}

Cross-dataset deepfake detection remains difficult because whole-face
detectors often learn the wrong evidence.
A model trained on FaceForensics++~\cite{rossler2019faceforensics} can
achieve high in-distribution accuracy while relying on signals that do
not transfer: codec traces, compression level, background statistics,
actor distribution, or manipulation-method-specific texture artifacts.
When evaluated on Celeb-DF v2~\cite{li2020celebdf} or
DFD~\cite{dufour2019dfd}, the facial manipulation signal may remain,
but many of the dataset-specific shortcuts change.
The central problem is therefore not only which backbone to use, but
how the facial evidence is presented to the classifier.

Most deepfake detectors aggregate first and explain later.
Whole-face CNNs pool spatial evidence into a global descriptor before
classification.
Vision-transformer detectors often use a CLS token that compresses
the full face into a single representation.
Even when overlaid saliency maps are applied, the classifier has
already seen the whole image, including regions that may carry
dataset-specific shortcuts.
This order of operations is problematic for forensic use: pooling can
dilute localized manipulation evidence, while overlaid explanations
cannot guarantee which evidence the classifier actually used.

We propose the opposite design order:
\textbf{segment first, filter second, pool third, classify last}.
A frozen FaRL~\cite{zheng2022farl} face parser assigns each DINOv3
ViT-L/16~\cite{oquab2024dinov3} patch token to a semantic face region.
Tokens outside the selected region are discarded before aggregation,
and only the remaining regional tokens are mean-pooled and passed to
a linear probe.
In this design, segmentation is not used as a visualization tool or
as a preprocessing crop.
It becomes a classification operator: it defines the input space of
the detector before the decision is made.

This produces a compact forensic package.
First, the system detects whether a face is real or fake.
Second, it localizes the semantic source of the evidence because the
classifier is restricted to a known region, such as the mouth.
Third, the attribution is structural: if the mouth-indexed model
predicts fake, the decision used only mouth tokens, not inferred from
an overlaid saliency map.
The explanation is built into the model's input pathway, not
reconstructed after classification.

The key technical claim is that the order of spatial operations
matters.
Pooling all patch tokens before classification mixes
manipulation-relevant evidence with irrelevant facial, background,
and acquisition cues.
By contrast, filtering tokens by semantic region before pooling
preserves the semantic origin of the signal and reduces the
classifier's exposure to region-irrelevant shortcuts.
This is especially important for frozen vision foundation models,
whose dense patch tokens already contain spatially structured
information.
DINOv3 provides a spatially consistent patch-token grid; FaRL
provides the semantic index that turns this grid into
region-specific token subsets.

Our experiments show that this simple design choice has a large
effect on cross-dataset generalization.
A linear probe trained on mouth-indexed frozen DINOv3 tokens achieves
AUC 0.905 on Celeb-DF v2 and 0.930 on DFD, with no DINOv3 or FaRL
fine-tuning and no target-domain data.
Bypassing the spatial index and using DINOv3's CLS token on the same
face crop drops Celeb-DF v2 AUC to 0.641, while replacing DINOv3
with FaRL features drops it to 0.696.
These ablations show that neither segmentation alone nor the frozen
backbone alone explains the result; the gain comes from their
composition through pre-pooling spatial indexing.

\vspace{2pt}\noindent\textbf{Contributions.}
\begin{itemize}\setlength\itemsep{0pt}
  \item \textbf{Segmentation as a classification enabler.}
    We introduce segmentation-guided spatial indexing, where a frozen
    face parser converts dense DINOv3 patch tokens into
    region-specific token subsets before pooling.
  \item \textbf{Filter-before-pool design.}
    We show that semantic filtering before aggregation preserves
    manipulation-relevant regional evidence that is diluted by
    CLS-token or whole-face pooling.
  \item \textbf{One-pass forensic interpretability.}
    The same forward pass produces a fake probability, region-level
    evidence localization, and a structural explanation of what
    semantic evidence the classifier was allowed to use.
  \item \textbf{Frozen, target-free generalization.}
    The system uses a frozen DINOv3 backbone, a frozen FaRL parser,
    and a linear probe, with no DINOv3 or FaRL fine-tuning and no
    target-domain data.
    On Celeb-DF v2 and DFD, it achieves strong cross-dataset
    performance in a frozen-backbone, region-restricted probing regime.
\end{itemize}

\section{Related Work}
\label{sec:related}

Deepfake detectors often perform well within a benchmark but degrade
under cross-dataset evaluation.
Models trained on FaceForensics++~\cite{rossler2019faceforensics} can
learn manipulation evidence together with dataset-specific shortcuts
such as compression level, codec traces, and actor distribution,
making high in-distribution performance an unreliable indicator of
forensic robustness.
Xception~\cite{rossler2019faceforensics}, F3-Net~\cite{zhao2021f3net},
and RECCE~\cite{cao2022recce} improve supervised detection within
controlled settings, but their end-to-end training allows the model
to exploit dataset-specific cues that do not transfer to Celeb-DF v2 or DFD.
Our work addresses this by restricting the classifier to semantically
selected regions before pooling, reducing exposure to whole-face
shortcuts while preserving localized manipulation evidence.

Several methods exploit the non-uniform spatial distribution of facial
manipulations.
LipForensics~\cite{haliassos2021lipforensics} focuses on the mouth
through lipreading-based pretraining and achieves strong cross-dataset
performance, showing that localized evidence can generalize better
than whole-face appearance.
Face2Parts~\cite{uddin2026face2parts} decomposes features by facial
region using channel attention and deep triplet learning, achieving
CelebDF-v2 0.853, the closest prior result to ours structurally,
but with a fine-tuned backbone and learned aggregation.
In our framework, segmentation is a classification operator: it
indexes frozen DINOv3 patch tokens before pooling and defines the
classifier's input space; non-target tokens are removed before the
descriptor is formed.

Frozen-feature detectors such as
UniversalFakeDetect~\cite{ojha2023towards} and the LayerNorm-tuned
CLIP probe of Yermakov et al.~\cite{yermakov2025unlocking} show that
large pretrained representations can resist dataset-specific
overfitting.
However, these methods rely on global image representations or
CLS tokens and do not provide semantic evidence localization by
construction.
Our method instead exploits DINOv3's dense patch-token grid: a frozen
face parser converts it into semantic subsets, making the detector
both frozen-backbone and region-restricted.

DINO~\cite{caron2021dino} established emergent spatial structure in
self-supervised ViTs; DINOv3~\cite{oquab2024dinov3} attributes this
to Gram training, which enforces consistent feature geometry across
all patch positions and enables both dense prediction and, as we show,
spatial filtering for classification.
FaRL~\cite{zheng2022farl} provides the semantic face-part index:
the parser assigns semantic labels, DINOv3 supplies region-specific
patch features, and the classifier reads only the selected regional
descriptor: segment, filter, pool, classify.

Most explainability methods are retrospective: saliency maps,
Grad-CAM, and attention heatmaps are applied after the model has
already processed the full face, and cannot guarantee that the
classifier relied only on the highlighted region.
Our framework provides structural interpretability: the classifier
is restricted to a known semantic region before inference, so if the
mouth-indexed detector predicts fake, the decision was made from
mouth tokens only, making the input pathway auditable by construction.
To the best of our knowledge, this is the first deepfake-detection
framework to use semantic face segmentation as a pre-pooling spatial
index over frozen DINOv3 patch tokens.

\section{Method}
\label{sec:method}

\begin{figure}[h!]
  \centering
  \includegraphics[width=\columnwidth]{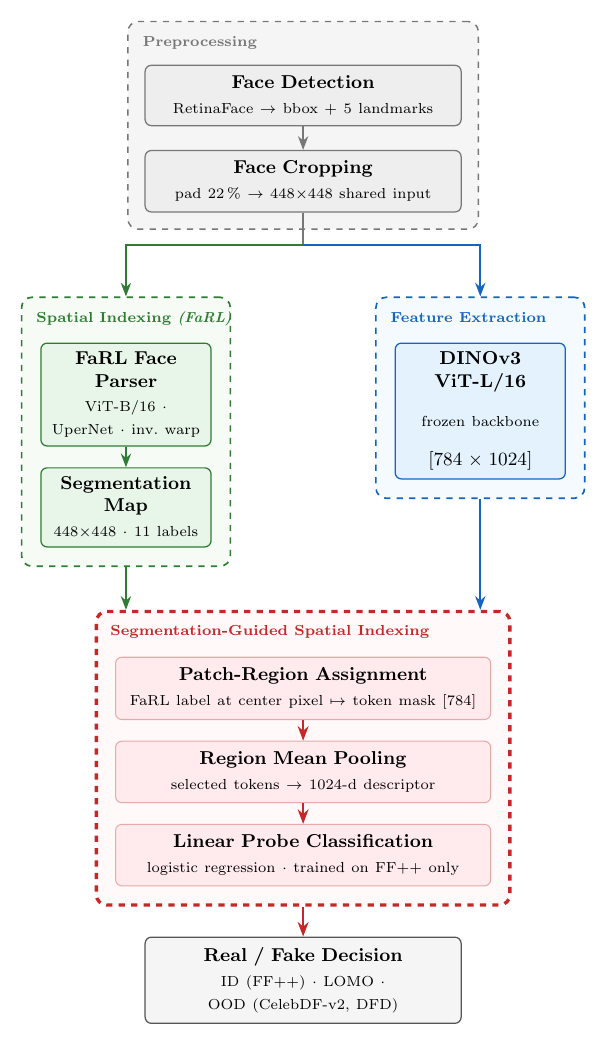}
  \caption{Segmentation-guided region selection framework. FaRL assigns
    each of DINOv3's 784 frozen patch tokens a semantic face-part
    label; mean pooling within the target region yields a 1024-d
    descriptor for a linear probe.}
  \label{fig:framework}
\end{figure}

Standard face detectors first compress the face into a global
representation, through a CNN pooling layer, a transformer CLS token,
or whole-face patch pooling, and only then ask a classifier to separate
real from fake.
At that point, localized forensic evidence has already been mixed with
identity, skin texture, compression traces, background, and other
dataset-specific cues.

We reverse this order.
Before any regional descriptor is formed, a frozen face parser assigns
each DINOv3 patch token to a semantic face part.
Tokens outside the selected region are removed, and only the remaining
region-specific tokens are pooled and passed to a linear probe.
Segmentation therefore becomes part of the classifier's input pathway:
it decides what evidence the classifier is allowed to see before
classification happens.

This produces a region-restricted detector from two frozen components.
DINOv3 provides the dense patch-token representation; FaRL provides
the semantic index over that token grid.
Their composition turns the $28{\times}28$ patch grid into semantic
token subsets such as mouth, eyes, nose, brows, skin, or hair.
A detector trained on the mouth subset is not a whole-face model later
explained by a saliency map; it is a model whose input was restricted
to mouth tokens by design.

\subsection{Preprocessing}

For each video frame, we detect the face with
RetinaFace~\cite{deng2020retinaface} and crop it with 22\% padding.
The crop is resized to $448{\times}448$ and used for both branches:
the DINOv3 feature branch and the FaRL segmentation branch.

The $448{\times}448$ resolution is important because DINOv3 ViT-L/16
produces one patch token per $16{\times}16$ image patch, yielding a
$28{\times}28$ grid of 784~spatially addressable tokens.
This density ensures that small facial regions such as mouth, eyes,
nose, and brows are each represented by multiple tokens rather than
collapsed into a single global descriptor.

The same face crop is used for all ablations, so comparisons between
CLS, whole-face patch pooling, and segmentation-guided indexing are
not confounded by face detection or image resolution.

\subsection{Spatial Indexing (FaRL) and DINOv3 Feature Extraction}

The preprocessed crop is passed through two frozen branches.

\noindent\textbf{FaRL branch.}
RetinaFace landmarks drive a similarity transform $+$ tanh-warp into
canonical pose; a frozen FaRL ViT-B/16$+$UperNet parser predicts a
pixel-level semantic segmentation map over face parts; the
segmentation is mapped back to crop coordinates.

\noindent\textbf{DINOv3 branch.}
The same $448{\times}448$ crop is forwarded through frozen DINOv3
ViT-L/16.
Block-20 activations are extracted (\texttt{get\_intermediate\_layers},
\texttt{norm=False}), yielding 784 patch tokens
$\in\mathbb{R}^{1024}$.
The backbone is never fine-tuned.
Patch features are standardised per embedding dimension using
statistics from the FF++ training split only; no Celeb-DF~v2 or DFD
statistics are used.

FaRL provides semantic addressability: which face part each patch
belongs to.
DINOv3 provides the transferable dense representation: the patch-level
features used for classification.
The method depends on their composition, not on either branch alone.

\subsection{Segmentation-Guided Spatial Indexing}

Each $16{\times}16$ DINOv3 patch is assigned the FaRL label of its
center pixel: for patch $(i,j)$, we read the label at pixel
$(16i{+}8,\;16j{+}8)$, producing a 784-element semantic index
aligned with the 784 patch tokens.

Given a target region, we retain only the tokens whose FaRL label
belongs to that region; all others are discarded before aggregation.
Formally, let $X \in \mathbb{R}^{784 \times 1024}$ be the patch-token
matrix and $s \in \{1,\ldots,K\}^{784}$ the label vector.
A whole-face detector forms $z_{\text{whole}} = \mathrm{Pool}(X_1,
\ldots, X_{784})$; a CLS token aggregates the full face globally.
Our method instead applies: $z_r = \mathrm{Pool}(\{X_i : s_i = r\})$.
The classifier receives $z_r$; its decision boundary is learned only
over features from the selected semantic region.
Selected tokens are mean-pooled to a 1024-d descriptor; an
$\ell_2$-regularised logistic regression ($C{=}0.1$, class-balanced)
trained on FF++ provides the binary classifier; frame-level
probabilities are averaged to a video-level score.

This filtering step is the central mechanism.
The model is structurally restricted to the target region before the
descriptor is formed.
The same forward pass therefore provides detection, region-level
evidence localization, and structural attribution: when the
mouth-indexed model predicts fake, the decision was made only from
mouth tokens.

\begin{figure}[h!]
  \centering
  \includegraphics[width=\linewidth]{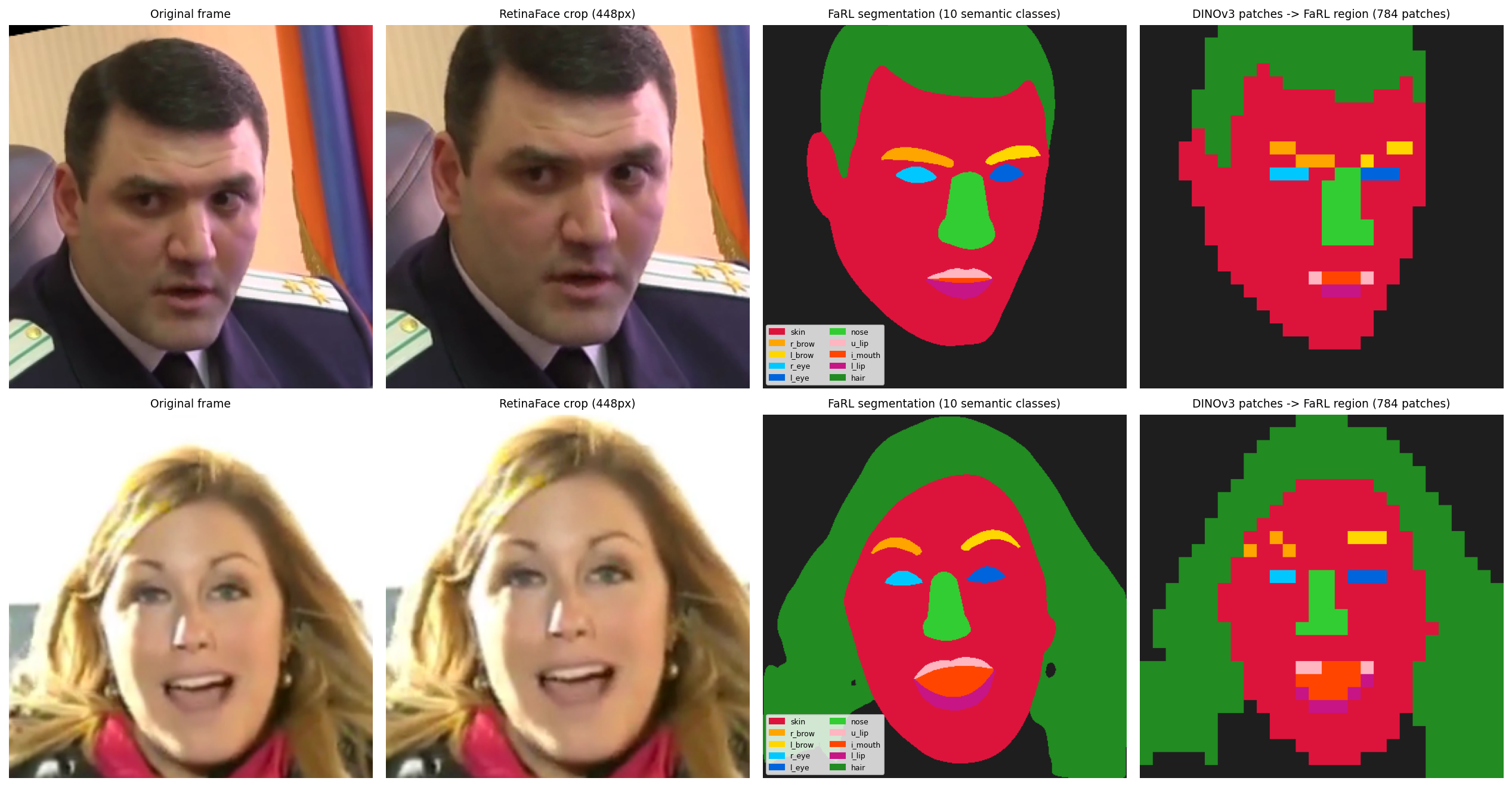}
  \caption{FaRL segmentation aligned to the DINOv3 patch grid.
    Left to right: face crop, FaRL pixel labels, patch tokens
    color-coded by region assignment via center-pixel lookup.}
  \label{fig:demo}
\end{figure}

\section{Experiments}
\label{sec:experiments}

The experiments are organized around the central claim: for localized
forensic evidence, the order of operations matters.
Pooling the whole face first mixes manipulation cues with identity,
texture, compression, and other dataset-specific signals.
Filtering the token grid first gives the classifier a descriptor whose
semantic source is fixed before training and inference.

We evaluate not only whether the method performs well, but whether
performance follows from the proposed mechanism.
Four questions guide the experiments: Does filtering before pooling
outperform global CLS or whole-face representations?
Is the gain from segmentation alone, DINOv3 alone, or their
composition?
Which facial regions transfer beyond FF++?
Where does static region-restricted detection break down?

\noindent\textbf{Datasets and protocol.}
FaceForensics++~\cite{rossler2019faceforensics}: 999 source videos,
four manipulation types (DF, F2F, FS, NT), c23 compression, 15
frames/video.
Celeb-DF~v2~\cite{li2020celebdf}: 518 real, 5639 fake videos.
DFD~\cite{dufour2019dfd}: 363 real/fake pairs.
All probes are trained on FF++ and evaluated cross-dataset on
Celeb-DF~v2 and DFD; no target-domain labels are used.
The main configuration uses DINOv3 ViT-L/16 block-20 patch features,
FaRL semantic indexing, mouth-token selection, mean pooling, and a
linear probe.

\subsection{Does filtering before pooling matter?}

\begin{table}[t]
\centering
\caption{Mechanism ablation (Mouth region). All models use the same
  face crop. \textbf{Bold}: best column.}
\label{tab:ablation}
\resizebox{\columnwidth}{!}{%
{\small
\begin{tabular}{llccc}
\toprule
Model & Filter & Features & CelebDF v2 & DFD \\
\midrule
DINOv3 CLS            & No  & CLS token & 0.641 & 0.805 \\
Whole-face patch pool & No  & DINOv3    & 0.838 & 0.889 \\
FaRL Mouth            & Yes & FaRL      & 0.696 & 0.811 \\
Ours (final layer)    & Yes & DINOv3    & 0.833 & 0.900 \\
\textbf{Ours (blk\,20)} & \textbf{Yes} & \textbf{DINOv3}
                      & \textbf{0.905} & \textbf{0.930} \\
\bottomrule
\end{tabular}%
}
}
\end{table}

Replacing mouth-selected tokens with DINOv3's CLS token drops
CelebDF~v2 AUC from 0.905 to 0.641 ($-$26.4\,pp).
Whole-face patch pooling is stronger than CLS, but still 6.7\,pp
below mouth-indexed pooling: the decisive step is semantic filtering
before aggregation, not merely using patch tokens over CLS.
The FaRL-mouth baseline uses the same region but replaces DINOv3
with FaRL features, reaching only 0.696.
Segmentation alone is not enough; the gain comes from composing
FaRL's semantic index with DINOv3's transferable dense representation.

\subsection{Which regions transfer beyond FF++?}

\begin{table}[t]
\centering
\caption{Per-region AUC: FF++ ID and zero-shot OOD.
  \textbf{Bold}: best per column.}
\label{tab:region_matrix}
\setlength{\tabcolsep}{5pt}
{\small
\begin{tabular}{lccc}
\toprule
Region     & FF++  & CelebDF v2      & DFD   \\
\midrule
Mouth      & 0.982 & \textbf{0.905}  & \textbf{0.930} \\
Nose       & 0.977 & 0.870           & 0.887 \\
Skin       & 0.978 & 0.857           & 0.900 \\
Eyes       & 0.976 & 0.837           & 0.890 \\
Brows      & \textbf{0.989} & 0.817  & 0.902 \\
Hair       & 0.914 & 0.756           & 0.820 \\
\midrule
Whole face & 0.967 & 0.838           & 0.889 \\
\bottomrule
\end{tabular}
}
\end{table}

The region ranking changes sharply between in-distribution and
cross-dataset evaluation.
Brows achieves the best FF++ AUC (0.989) but collapses to 0.817 OOD,
the weakest non-Hair result.
Mouth is not the top FF++ region, yet leads on both OOD benchmarks.
A region chosen by FF++ performance alone would select Brows and
incur a $-$8.8\,pp penalty on CelebDF~v2.
Hair provides a negative control: no FF++ manipulation type alters
hair, so a hair-restricted probe learns dataset-specific texture that
does not transfer.

\subsection{How does the method compare with prior detectors?}

\begin{table}[t]
\centering
\caption{Cross-dataset OOD comparison. All methods trained on FF++.
  \textbf{Bold}: best column.}
\label{tab:sota}
\setlength{\tabcolsep}{4pt}
{\small
\begin{tabular}{llcc}
\toprule
Method & Backbone & CelebDF v2 & DFD \\
\midrule
Xception~\cite{rossler2019faceforensics}      & Fine-tuned    & 0.736 & 0.816 \\
LipForensics~\cite{haliassos2021lipforensics} & Task-specific & 0.824 & ---   \\
Face2Parts~\cite{uddin2026face2parts}         & Fine-tuned    & 0.853 & 0.894 \\
\midrule
Ours (Nose)              & Frozen        & 0.870 & 0.887 \\
\textbf{Ours (Mouth)}    & \textbf{Frozen} & \textbf{0.905} & \textbf{0.930} \\
\bottomrule
\end{tabular}
}
\end{table}

Mouth outperforms LipForensics by $+$8.1\,pp and Xception by
$+$16.9\,pp on CelebDF~v2, and Xception by $+$11.4\,pp on DFD,
with a frozen backbone and no forgery-specific fine-tuning
(Table~\ref{tab:sota}).
Face2Parts also decomposes by facial region yet reaches only 0.853
on CelebDF~v2 ($-$5.2\,pp), despite fine-tuning its backbone with
channel attention and deep triplet learning.
Methods achieving higher absolute AUC~\cite{yermakov2025unlocking}
do so by updating backbone parameters or training a side network;
our claim is that segmentation-guided indexing turns frozen dense
DINOv3 features into a strong cross-dataset detector with an explicit
evidence pathway.

\subsection{Does region evidence follow the manipulation type?}

\begin{table}[h]
\centering
\caption{LOMO AUC by region (block-20). NT collapses across all
  regions, a structural limit of static semantic indexing.}
\label{tab:lomo}
{\small
\begin{tabular}{lccc}
\toprule
Held-out & Eyes & Mouth & Nose \\
\midrule
DeepFakes      & 0.954 & 0.929 & 0.948 \\
Face2Face      & 0.744 & 0.847 & 0.808 \\
FaceSwap       & 0.771 & 0.626 & 0.693 \\
NeuralTextures & 0.457 & 0.571 & 0.538 \\
\bottomrule
\end{tabular}
}
\end{table}

Eyes transfer best for DeepFakes (0.954), consistent with
identity-swap blending seams at the periocular boundary; Mouth leads
Face2Face (0.847), consistent with expression reenactment at the
lips.
The spatial index exposes region-specific forensic evidence aligned
with the manipulation locus.
NeuralTextures collapses across all regions (0.457--0.571): static
semantic indexing is weak when manipulation is globally diffuse
rather than localized, a structural limit of any static region-based
detector.

\subsection{Where in DINOv3 is regional evidence strongest?}

\begin{table}[h]
\centering
\caption{OOD AUC by DINOv3 block (Mouth region).
  Final layer is weakest due to normalization mismatch.}
\label{tab:block_sweep}
{\small
\begin{tabular}{lcc}
\toprule
Block       & CelebDF v2 & DFD   \\
\midrule
Block 16    & 0.869      & 0.911 \\
Block 18    & 0.881      & 0.930 \\
\textbf{Block 20}  & \textbf{0.905} & 0.930 \\
Block 22    & 0.904      & \textbf{0.934} \\
Final layer & 0.833      & 0.900 \\
\bottomrule
\end{tabular}
}
\end{table}

Intermediate blocks 20--22 are near-optimal; the final layer is
weakest on both benchmarks (7.2\,pp below block~20 on CelebDF~v2),
reflecting the normalization mismatch described in
\S\ref{sec:method}.
Block~20 is the primary configuration; block~22 produces comparable
performance.

For deployment: mouth-indexed DINOv3 tokens are the recommended
configuration when a parser is available; whole-face patch pooling
is the correct fallback without parsing ($-$6.7\,pp on CelebDF~v2).
Region selection by in-distribution AUC alone favours Brows and
costs $-$8.8\,pp OOD relative to Mouth.

\section{Discussion}
\label{sec:discussion}

\noindent\textbf{Segmentation as an access key to dense VFM geometry.}
The role of FaRL in this framework is different from how face parsing
is usually used in deepfake detection.
In region-aware methods, a parsing map often defines where a
specialized detector should crop, attend, or apply auxiliary
supervision.
In our method, the parsing map is an access key into the dense DINOv3
feature grid: it determines which patch tokens in the $784{\times}1024$
matrix reach the probe.
This reframes segmentation as a classification operator.
The face parser defines the detector's input before classification,
not after.
A mouth-indexed model is therefore not a whole-face model with a mouth
saliency map; it is a model whose descriptor was formed only from
mouth-associated DINOv3 tokens.
This mechanism works because DINOv3 patch tokens are spatially
addressable: FaRL provides the semantic index, DINOv3 provides the
transferable patch-level representation, and the two frozen components
compose.

\noindent\textbf{Why this matters for in-the-wild forensics.}
Freezing DINOv3 and FaRL prevents FF++ supervision from reshaping
either feature extractor toward benchmark-specific artifacts.
The spatial index makes the evidence pathway explicit: if the mouth
model returns a high fake probability, the classifier was shown only
mouth tokens, a qualitative difference from Grad-CAM where the model
has already processed the full face.
Together, these properties address the two axes that benchmark AUC
alone hides: robustness under distribution shift and accountability of
the evidence pathway.

\noindent\textbf{What the region results reveal.}
The region-ranking inversion is a central finding: Brows achieves the
highest FF++ AUC but Mouth transfers better OOD.
In-distribution region performance is not a reliable proxy for
transferable forensic evidence.
Without region-restricted probes, this behavior would be hidden inside
a whole-face descriptor; the spatial index is also a diagnostic tool
for understanding where source-domain semantic evidence is
transferable.

\noindent\textbf{Scope and forward direction.}
NeuralTextures marks the boundary of static region indexing: when
manipulation is globally diffuse rather than localized, no single face
part carries reliable evidence.
The LOMO results show that different manipulation families leave
evidence in different regions, which points toward adaptive routing
over the same frozen, spatially indexed representation.
The broader design principle is that dense VFM features should not
always be collapsed into a global descriptor: \textbf{segmentation can
be the gate through which frozen VFM evidence reaches the
classifier.}

\section{Conclusion}
\label{sec:conclusion}

We introduce segmentation-guided spatial indexing as a
filter-before-pooling design for frozen-VFM deepfake detection.
Instead of forming a whole-face descriptor first, the method uses a
frozen FaRL parser to assign DINOv3 patch tokens to semantic regions,
removes non-target tokens before aggregation, and trains a linear
probe on the resulting regional descriptor.

The main finding is that the order of operations matters.
When the classifier receives mouth-indexed DINOv3 tokens, it reaches
AUC 0.905 on Celeb-DF v2 and 0.930 on DFD without DINOv3 fine-tuning
or target-domain data.
When the same frozen backbone and crop are represented by the CLS
token, Celeb-DF v2 AUC drops to 0.641.
The gain comes not only from the backbone, but from the spatial index
that controls what evidence reaches the classifier before pooling.

The method also provides region-level evidence localization by
construction: a mouth-indexed detector makes its decision from mouth
tokens only, built into the input pathway rather than inferred
afterward through saliency.
The result is a simple forensic package (detection, semantic evidence
localization, and structural interpretability) and a design principle:
\textbf{filter before pooling}.

\smallskip\noindent\textbf{Limitations.}
NeuralTextures-style globally diffuse manipulations are not detected
by any static region (near-chance in LOMO).
Mean pooling discards intra-region spatial structure; attention-based
aggregation over the same selected tokens is the natural extension.
Static region selection cannot adapt to the manipulation type at test
time; per-sample adaptive routing is the subject of follow-on work.

\smallskip\noindent\textit{Ethics.}
This work uses face datasets containing identifiable individuals.
The method fails on global manipulations such as NeuralTextures and
has not been audited across demographic subgroups; any operational use
requires application-specific fairness and privacy review.

\bibliographystyle{IEEEbib}
\bibliography{refs,strings}

\end{document}